\documentclass{article}

\usepackage{microtype}
\usepackage{graphicx}
\usepackage{longtable}
\usepackage{subfigure}
\usepackage{booktabs} 
\usepackage{enumitem}
\usepackage[table]{xcolor}
 
\newcommand{\smallsec}[1]{\vspace{1pt} \noindent \textbf{#1.}}
\newcommand{\smallsecemph}[1]{\vspace{1pt} \noindent \emph{#1.}}

\usepackage{hyperref}

\usepackage[accepted]{icml2024}

\usepackage{amsmath}
\usepackage{amssymb}
\usepackage{mathtools}
\usepackage{amsthm}

\usepackage[capitalize,noabbrev]{cleveref}

\theoremstyle{plain}

\theoremstyle{definition}

\theoremstyle{remark}

\usepackage{soul}
\usepackage{tabularx}

\usepackage[textsize=tiny]{todonotes}

\icmltitlerunning{Measure Dataset Diversity, Don't Just Claim It}

\begin{document}

\twocolumn[
\icmltitle{Position: Measure Dataset Diversity, Don't Just Claim It}



\icmlsetsymbol{workdone}{*}

\begin{icmlauthorlist}
\icmlauthor{Dora Zhao}{workdone,yyy}
\icmlauthor{Jerone T. A. Andrews}{comp}
\icmlauthor{Orestis Papakyriakopoulos}{workdone,sch}
\icmlauthor{Alice Xiang}{comp2}
\end{icmlauthorlist}

\icmlaffiliation{yyy}{Stanford University, Stanford, USA}
\icmlaffiliation{comp}{Sony AI, London, UK}
\icmlaffiliation{comp2}{Sony AI, Seattle, USA}
\icmlaffiliation{sch}{Technical University of Munich, Munich, Germany}

\icmlcorrespondingauthor{Dorothy Zhao}{dorothyz@stanford.edu}

\icmlkeywords{Machine Learning, ICML}

\vskip 0.3in
]



\printAffiliationsAndNotice{\textsuperscript{*}Work done at Sony AI }  

\begin{abstract}
Machine learning (ML) datasets, often perceived as neutral, inherently encapsulate abstract and disputed social constructs. Dataset curators frequently employ value-laden terms such as diversity, bias, and quality to characterize datasets. Despite their prevalence, these terms lack clear definitions and validation. Our research explores the implications of this issue by analyzing ``diversity'' across 135 image and text datasets. Drawing from social sciences, we apply principles from measurement theory to identify considerations and offer recommendations for conceptualizing, operationalizing, and evaluating diversity in datasets.  Our findings have broader implications for ML research, advocating for a more nuanced and precise approach to handling value-laden properties in dataset construction.
\end{abstract}

\section{Introduction}
\label{sec:introduction}
Cloaked under the guise of objectivity, machine learning (ML) datasets are portrayed as impartial entities, giving the illusion of reflecting an ``unbiased look'' at the world~\citep{torralba2011unbiased}. Yet, beneath this veneer, datasets are not neutral---they are infused with values, bearing the indelible imprints of social, political, and ethical ideologies woven into their fabric by their curators~\citep{raji2ai,blili2023making,maleve2021data}. 

This inherent value-laden nature becomes glaringly apparent in the perpetuation of social stereotypes and the stark underrepresentation of marginalized communities within the lifecycle of ML datasets~\citep{wang2022revise,buolamwini2018gender,Zhao_2021,birhane2021multimodal,denton2020bringing}. From inception to release, datasets emerge as political artifacts, etched with the signature of their creators' perspectives, organizational priorities, and the broader cultural zeitgeist, making them potent instruments in shaping narratives and reinforcing power structures~\citep{winner2017artifacts,hanna2020against,birhane2022values}.

This politicization of datasets is particularly conspicuous in the criteria set by curators. Terms related to diversity, bias, quality, realism, difficulty, and comprehensiveness are frequently invoked~\citep{scheuerman2021datasets}, despite a glaring lack of consensus regarding their precise definitions. For instance, diversity dimensions can encompass a multitude of concepts, spanning ``dressing styles''~\citep{Bai_2021}, ``weather''~\citep{diaz2022ithaca365}, and ``ethnic[ity]''~\citep{Fu_2021} to ``verbs''~\citep{Sadhu_2021}, ``sentential contexts''~\citep{Culkin_2021}, and ``conversation forms''~\citep{Fabbri_2021}. Diversity can also refer to part of the collection process, such as recruiting annotators with ``diversity in gender,
age, occupation/background (linguistic and ethnographic knowledge), region (spoken dialects)''~\citep{zeinert2021annotating}, and ``psychological personality''~\citep{Chawla_2021}.

Recognizing this ambiguity, the need for precise and unambiguous definitions becomes paramount to ascertain whether datasets genuinely embody the proclaimed qualities. Treating value-laden constructs, such as diversity, bias, and quality, as self-evident perpetuates the fallacious belief that datasets are inherently neutral. Instead, we posit that datasets serve as tools wielded by curators to quantify abstract social constructs. Interrogating these values demands critical questions: \emph{How are these constructs defined and operationalized? And how do we validate that datasets genuinely encapsulate the values they claim to represent?}

In this position paper, we leverage \emph{measurement theory}, a framework widely employed in the social sciences, to develop precise numerical representations of abstract constructs~\citep{bandalos2018measurement}. This application is integral to our focused analysis of \emph{diversity}---a frequently touted trait in ML datasets~\citep{scheuerman2019computers}---providing a structured approach to conceptualizing, operationalizing, and evaluating claimed dataset qualities. Our scrutiny extends to 135 text and image datasets, where we uncover key considerations and offer recommendations for applying measurement theory to their collection. We underscore the imperative for transparency in articulating how diversity is defined (\Cref{sec:conceptualization}) and how the data collection process aligns with this definition (\Cref{sec:operationalize}). Further, we present methodologies for evaluating diversity, scrutinized through the lenses of reliability (\Cref{sec:reliability}) and validity (\Cref{sec:validity}). Our work culminates with a case study on the Segment Anything dataset~\citep{kirillov2023segment}, illustrating the practical application of our recommendations (\Cref{sec:case_study}). 

\textbf{In summary, this position paper advocates for clearer definitions and stronger validation methods in creating diverse datasets, proposing measurement theory as a scaffolding framework for this process.}

\begin{figure*}[htbp]
    \centering
    \includegraphics[width=\textwidth, trim = 0cm 0cm 0cm 0cm, clip]{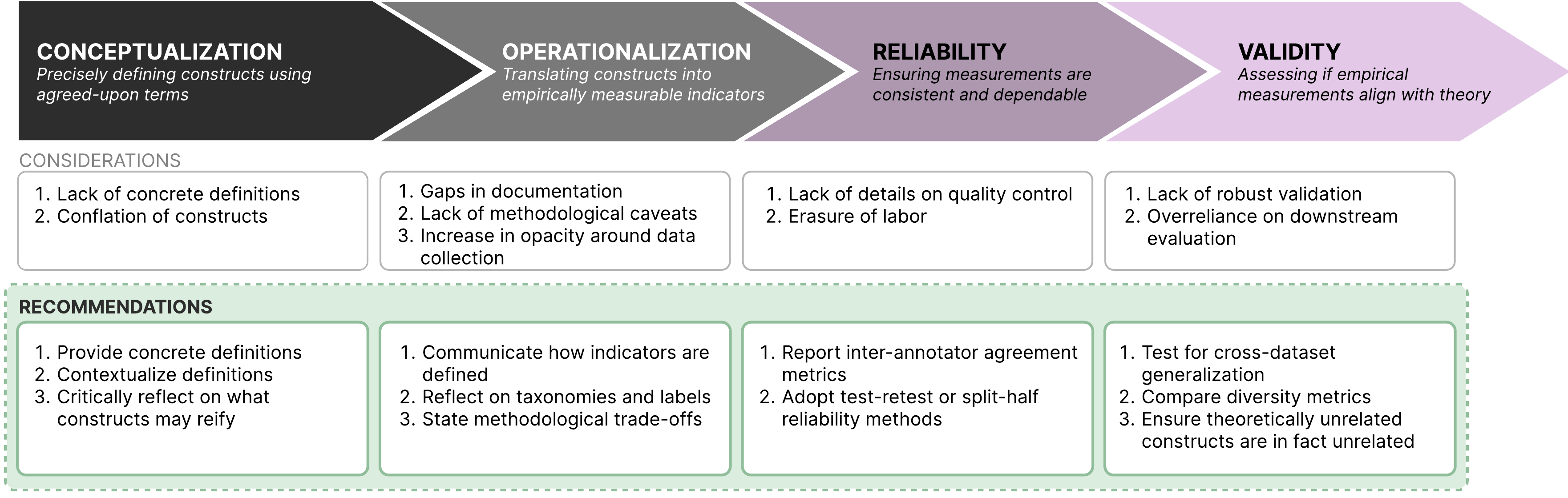}
    \caption{Overview of considerations and recommendations for  conceptualizing (\Cref{sec:conceptualization}), operationalizing (\Cref{sec:operationalize}), and evaluating the reliability (\Cref{sec:reliability}) and validity (\Cref{sec:validity}) of diverse datasets.}
    \label{fig:overview}
\end{figure*}

\section{Background} 
\label{sec:background}
\emph{Measurement} is a fundamental aspect of ML systems~\citep{jacobs2021measurement,jacobs2021governance}. These systems employ observable, real-world instances to quantify abstract constructs, including ``moral foundations''~\citep{Johnson_2018}, ``emotion''~\citep{wei2020learning}, and ``gender''~\citep{Wang_2019_Racial}. However, given the unobservable nature of these constructs, researchers rely on proxies and inference. For instance, the identification and prevalence of specific linguistic features, like derogatory language, serve as proxies to deduce the existence of misogynistic content in text~\citep{Zeinert_2021}. Assessing the quality of resulting datasets requires careful consideration of the validity of chosen proxies and the reliability of assumptions underlying ML systems. This is crucial, as proxies may normalize inadequacies without acknowledging limitations~\citep{andrus2021we}. Such considerations lead us to measurement theory in the social sciences, which offers methodologies for quantifying and encapsulating theoretical constructs that resist direct measurement~\citep{bandalos2018measurement}.

Conceptually, measurement theory provides a structured approach to move from latent, abstract constructs to observable, real-world variables. \emph{Conceptualization} involves precisely defining constructs using agreed-upon terms. Researchers then \textit{operationalize} these concepts by translating them into observable indicators that can be empirically measured in the real world~\citep{babbie2020practice, check2011research}. Finally, the measurements undergo \emph{evaluation}, considering \emph{reliability} and \emph{validity}~\citep{bandalos2018measurement}.

Recent ML research explores the application of measurement modeling to refine the conceptualization and operationalization of constructs, including fairness~\citep{jacobs2021measurement}, bias~\citep{jacobs2020meaning} and intelligence~\citep{blili2023making}. This trend underscores a growing emphasis on leveraging measurement theory to refine evaluation metrics and benchmarks~\citep{xiao2023evaluating,zhou2022deconstructing,subramonian2023takes}. Principles from measurement theory have also been applied to datasets, where \citet{mitchell2022measuring} suggest measuring various facets to facilitate dataset curation and meaningful comparisons. 

Complementarily, we leverage measurement theory to enhance ML datasets by transforming implicit, value-laden properties into measurable constructs. In contrast to previous work~\citep{blodgett2021stereotyping} that concentrated on assessing the validity of natural language processing benchmark datasets to evaluate stereotyping, we adopt a different approach. Here, we view the data collection process \emph{itself} as the measurement to validate. This involves questioning the reliability of the collection process and exploring methods for researchers to validate claimed dataset properties.

Zooming in on the intricacies of diversity, we unearth inconsistencies in its definitions, a pervasive challenge that our paper addresses head-on. As in other fields with broad definitions of diversity~\cite{xu2020diversity}, there is no consensus on what diversity means for ML datasets. Rather than offering a prescriptive definition, we adopt the various interpretations provided by dataset collectors and develop an inductive taxonomy based on these definitions (see \Cref{tab:diversity_list}). 
While many ML datasets claim to be more diverse, neither what researchers mean by the term diversity nor how this property is achieved are clearly specified. Our aim, by shedding light on these disparities, is not only to guide dataset creators but also to equip reviewers with the necessary insights to critically evaluate authors' claims. Our work extends beyond the confines of dataset creation, presenting a broader contribution to the enhancement of ML and scientific practices. We summarize our considerations and recommendations in \Cref{fig:overview}.



\section{Method}
\label{sec:method}
To inform our position, we conducted a systematic literature review encompassing 135 datasets presented as being more ``diverse''. We choose to focus on image and text datasets, aligning with the prevailing emphasis on data collection practices in these domains~\citep{scheuerman2021datasets, paullada2021data, raji2ai}. Datasets were identified through searches across well-established venues in computer vision (ICCV, ECCV, CVPR), natural language processing (*CL), fairness (FAccT, AIES), and ML (NeurIPS, NeurIPS Datasets and Benchmarks Track), concluding with publications available until September 2022. Inclusion criteria involved retaining papers featuring both a diversity-related keyword (``divers*'', ``bias*'') and dataset-related keyword (``dataset'', ``annotation'') within their abstracts.

Our comprehensive analysis traverses the entire dataset collection pipeline, ranging from the motivation behind dataset creation to its eventual release and ongoing maintenance. To accomplish this, we consulted established guidelines for responsible data collection~\citep{andrews2023ethical,scheuerman2021datasets,peng2mitigating,google2019pair,gebru2021datasheets,holland2020dataset,bender2018data,pushkarna2022data,blodgett2022responsible}. The evolution of our codebook variables occurred iteratively during the coding process, driven by ongoing discussions among the authors. Refer to \Cref{sec:appendix_method} for more details.

\section{Conceptualization}
\label{sec:conceptualization}

Conceptualization plays a pivotal role in the research process, involving the definition and specification of clear constructs to be measured~\citep{check2011research,babbie2020practice}. For dataset creators, this phase resembles the translation of abstract values, such as diversity, into tangible and concrete definitions. Despite the widely acknowledged importance of diversity as a value for datasets~\citep{Van_Horn_2021,Yang_2021,Sugawara_2022,derczynski2016broad}, there is a notable lack of consistency and clarity among creators regarding its practical interpretation. This section examines conceptual challenges and provides recommendations.

\subsection{Considerations}
\smallsec{Lack of concrete definitions}
Recognizing the inherently abstract nature of diversity, a well-defined concept not only clarifies the significance of a diverse dataset but also lays the groundwork for operationalizing the collection process. Significantly, only 52.9\% of datasets explicitly justify the need for diverse data. For example,~\citet{rojas2022dollar} underscore the need for geodiverse data, driven by observations that previous datasets ``suffer from amerocentric and eurocentric representation bias that impacts the performance of classification tasks on images from other regions''. In this context, \emph{geodiversity} serves as a clearly defined construct, specifying the criteria for diversity based on geographical locations. Such constructs play a crucial role in streamlining the operationalization of data collection (e.g., employing photographers across 63 countries), facilitating a more straightforward validation process against the intended motivation. In essence, unambiguous definitions empower both consumers and reviewers to assess a dataset's suitability for specific tasks with heightened confidence. 

\smallsec{Conflation of constructs}
The lack of a standardized conceptualization approach can lead to unintended consequences, such as the paradoxical increase in offensive content observed in larger datasets~\citep{birhane2023hate}. This stems from misinterpretations where curators conflate \emph{scale} with diversity, assuming increased size inherently leads to more diverse data~\citep{bender2021dangers}. Scale may help in some contexts, such as achieving composition diversity, but relying solely on ``scale-thinking'' might overlook crucial nuances and idiosyncrasies inherent in specific tasks or setups~\citep{diaz2023scaling,hanna2020against}. 

There is also a tendency to conflate \emph{bias} with diversity~\citep{scheuerman2021datasets}, exemplified by attempts to mitigate bias in datasets for text summarization. As \citet{Li_2019} report, prior works ``alleviate [extractive] bias by collecting articles from diverse news publications.'' Establishing clear conceptual frameworks is not only a methodological necessity but also a crucial means to navigate complex relationships, recognizing that surface-level connections between diversity and other constructs, such as scale and bias, are not causally related.

\subsection{Recommendations}

\smallsec{Provide concrete definitions} 
Following the conceptualization phase, curators are urged to establish a precise definition of diversity, ensuring alignment and consistency throughout the dataset collection process~\citep{check2011research,babbie2020practice}. Emphasizing the importance of clear definitions, consider the subtle distinction between seemingly similar claims, such as representing ``diverse scenarios''~\citep{miao2022large} and ``diverse social scenes''~\citep{Fan_2018}. The former is defined in terms of things, stuff, and scene location (i.e., ``indoor and outdoor''), whereas the latter in terms of people, cultures, and scene location (i.e., ``living room, kitchen, restaurant, ...''). Definitional disagreements naturally arise, but providing an explicit definition signals the interpretation of diversity within the context of a dataset, enhancing assessment and enabling meaningful cross-dataset comparisons.

\smallsec{Contextualize definitions} 
When crafting definitions, curators must align interpretations with existing literature, evaluating theoretical underpinnings~\citep{blodgett2020language} and building on prior scholarship. In addressing \emph{skin tone} diversity, \citet{hazirbas2021towards} critically assess the drawbacks of ethnicity labeling practices in a previous facial attribute dataset~\citep{karkkainenfairface}, highlighting its subjectivity and potential to cause conceptual confusion. This scrutiny extends to citing ``unconscious biases'' linked to the ``other-race effect'' identified in psychology~\citep{o1996other}. In response, \citet{hazirbas2021towards} opt to annotate apparent skin tone, using the Fitzpatrick scale~\citep{fitzpatrick1975sun}, acknowledging limitations regarding the scale's reliability and validity~\citep{howard2021reliability}. Such contextualization plays a crucial role in positioning how a dataset \emph{contributes} to or \emph{challenges} prevailing notions.

\smallsec{Critically reflect on constructs} 
Definitions hold power. The choices we make in defining constructs, or even the decision to define certain constructs, bestow legitimacy upon specific beliefs. Thus, prior to advancing with the collection and release of a dataset, it is crucial to engage in a thoughtful reflection on the potential for \emph{reification}---the act of treating an abstract concept as something concrete~\citep{bhattacherjee2012social}. This tendency is frequently illustrated by the use of tangible test scores, such as those from IQ tests, to represent abstract concepts like ``intelligence''~\citep{blili2023making}. 

This concern gains particular significance in the realm of demographic diversity. For example, casting gender as binary reaffirms the normative notion of a binary gender system~\citep{hamidi2018gender}, and using racial categories may inadvertently endorse the fallacy that these groups are natural rather than social~\citep{hanna2020towards,khan2021one}.

Parallel considerations extend to text datasets. Nine surveyed text datasets~\citep{Orbach_2021,Abdul_Mageed_2021,Fabbri_2021,Peskov_2019,Durmus_2019,derczynski2016broad,ide2008masc,rahman2021information,Sugawara_2022} underscore \emph{source diversity} by drawing from various topic domains or websites. It is crucial to recognize that different source media inherently introduce their own values. Take, for example, the Broad Twitter Corpus~\citep{derczynski2016broad}, a diverse collection of English-speaking social media content sourced from the US, UK, New Zealand, Ireland, Canada, and Australia. Although the authors ``were constrained by
the number of local crowd workers'' accessible, the exclusion of English-speaking countries, ``such as Botswana and Singapore'', confines the dataset to a Western-centric view.

\section{Operationalization}
\label{sec:operationalize}

Operationalization involves the meticulous development of methodologies to empirically measure abstract concepts~\citep{check2011research,babbie2020practice}. In the context of ML datasets, this manifests as the tangible process of accumulating instances for a dataset. Within our corpus, we identify five primary dataset types categorizable by collection methodology: derivatives, ``real-world'' sampled, synthetically generated, web scraped, and crowdsourced (refer to \Cref{sec:appendix_dataset} for further details). In this section, we spotlight deficiencies in current collection processes and present recommendations to surmount these issues. 

\subsection{Considerations}
\label{subsec:operationalization_considerations}


\smallsec{Gaps in documentation} A significant concern centers on the insufficiency of information provided regarding datasets. Consistent with prior work~\citep{scheuerman2021datasets}, we observe that most papers introduce not only a dataset but also a new model, task, or algorithm. In particular, out of the 135 papers analyzed, only 40 are standalone dataset papers, or those where the dataset is positioned as the \emph{primary} contribution. Consequently, limited space is allocated for dataset creators to furnish detailed insights into collection strategies or the rationale behind methodological choices. This lack of (thorough) documentation results in unclear definitions from curators, creating uncertainty about the type of diversity sought and how these concepts are operationalized. For instance, among the 21 ``real-world'' sampled datasets, information on the location and time of data capture, as well as the identity of the data collectors (e.g., authors, researchers), is missing in 13 instances. Similarly, for web-scraped datasets, papers often omit collection criteria, such as the keyword search queries, crucial for understanding the dataset's sampled distribution and potential biases, be they social or methodological.

This \emph{documentation gap} is, in part, indicative of cultural attitudes prevalent in the ML research. As echoed previously, there exists a tendency to undervalue data work compared to model work or algorithmic contributions~\citep{sambasivan2021everyone}. Papers exclusively dedicated to datasets typically do not find their way into ``top-tier'' research venues~\citep{scheuerman2021datasets,Heinzerling_2022}, potentially dissuading authors from allocating sufficient time and space to elaborate on their dataset creation processes.




\smallsec{Lack of methodological caveats}
Even when datasets offer insights into their collection processes, methodological considerations or limitations are seldom addressed, accounting for 74.8\% of cases. Nonetheless, it is essential to acknowledge that every data collection method has inherent drawbacks. For example, web scraping offers a fast and cost-effective means to amass large quantities of data~\citep{ramaswamy2023beyond,raji2021face,li2023there}. However, concerns arise that this method might only capture ``canonical'' perspectives, particularly when utilizing Internet search engines~\citep{barbu2019objectnet,zhu2016object}. This issue may be exacerbated by, e.g., consensus-based quality filtering, which tends to \emph{exclude} noncanonical examples~\citep{mayo2022hard}. 

In contrast, synthetic generation, offering similar benefits to web scraping with increased systematic control~\citep{Johnson_2017,wood2021fake,Ros_2016,wei2020learning}, may introduce a domain gap between generated and real-world data~\citep{wei2020learning,Ros_2016}. Inclusion of these considerations is crucial as it justifies why the chosen method is best suited for capturing the proposed definition of diversity.

\smallsec{Increase in opacity}
We posit that the identified issues will worsen as data collection processes become more opaque. A clear manifestation of this trend is the increasing reliance on third-party assistants in data collection efforts~\citep{Orbach_2021,Dave_2020,miceli2022documenting}. While this approach may improve quality and efficiency, it introduces a layer of separation between those commissioning a dataset and those collecting it. This separation results in the loss of detailed knowledge, such as participant recruitment practices and quality assessment criteria, necessary for evaluating the validity of the collection process and the utility of the dataset. Furthermore, closed-source models and datasets raise additional concerns~\citep{bommasani2023foundation}. Examples include closed- and open-source models such as GPT-4~\citep{achiam2023gpt}, CLIP~\citep{radford2021learning}, Gemini~\citep{team2023gemini}, and Llama 2~\citep{touvron2023llama}. This growing opacity can lead to serious issues, as model consumers are unable to audit training datasets for potential biases or critique their construction methodology.

\subsection{Recommendations}

\smallsec{Clearly define variables of interest}
As highlighted by~\citet{raji2ai}, there is a prevailing inclination toward generality in ML research. This is evident in benchmarks claiming to evaluate ``general-purpose'' capabilities and datasets attempting to encompass diversity across numerous axes. However, just as it is impractical for a dataset to capture all the nuances and complexities of the world, it is equally implausible to collect diversity across every conceivable dimension~\citep{raji2ai}. For instance, defining diversity as ``variety across writing styles'' encompasses aspects such as, but not limited to, ``genre''~\citep{Soldan_2022}, ``narrative elements''~\citep{Xu_2022_Fairy}, or ``passage source, length, and readability''~\citep{Sugawara_2022}. 

Rather than pursuing generality, we advocate for curators to distinctly select and communicate specific variables that are most relevant to the task their dataset is meant to serve. Their definition of diversity, insights from prior datasets, considerations of the limitations in their collection practices, as well as rationale for excluding plausible aspects of diversity can inform this selection. For example, \citet{Van_Horn_2021} justify the decision to restrict the number of Animalia, Plantae and Fungi species in iNat2021 to those ``observed `enough' times by `enough' people'' during
a one-year time period, while transparently acknowledging the arbitrariness of their enumeration of ``enough''.



\smallsec{Communicate how indicators are defined} We encourage researchers to explicitly define the empirical indicators they use to measure diversity, specifying the scale, inclusion/exclusion criteria, and other relevant parameters. The variability in measurements across datasets makes it challenging to understand specific indicators without clear definitions. For example, Ithaca365~\citep{diaz2022ithaca365}, Mapillary Traffic Sign~\citep{Ertler_2020}, JHU-Crowd~\citep{sindagi2019pushing}, and MOTSynth~\citep{fabbri2021motsynth} all claim to include images representing diverse weather conditions. However, the measurement of weather diversity varies, ranging from employing nine categories (MOTSynth) to using only three (JHU-Crowd).



Clear definitions are paramount to prevent misinterpretations arising from diverse operationalizations, especially given that different operationalizations of the same construct can yield markedly different results. For example, indicators can be categorized as either \emph{objective} or \emph{subjective}. Objective indicators rely on explicit criteria and are assessed by external observers, while subjective indicators involve personal perception or evaluation. The YASO dataset~\citep{Orbach_2021} exemplifies this variability, where sentiment, typically considered subjective, is labeled by 7--10 crowdworkers. Although sentiment is inherently subjective, the YASO creators operationalize it as an objective measurement by reporting the majority sentiment among crowdworkers. Operationalizing sentiment as an objective measurement can obfuscate some of the nuances of sentiment that exist within the data. Such an approach fails to capture the full spectrum of sentiment, leading to limited understanding of sentiment diversity.


The approach by~\citet{Orbach_2021} also introduces potential concerns related to \emph{selection bias}. Under-performing workers are excluded based on ``test questions with an a priori known answer'' to ensure annotation quality. This practice may inadvertently filter out diverse perspectives or sentiments misaligned with the authors' subjective views, impacting the overall representativeness and objectivity of the YASO sentiment annotations. 

\smallsec{Critically reflect on taxonomies and labels}
Dataset curators should apply the same care during operationalization as in conceptualization, giving thoughtful consideration to label taxonomies, their suitability for the task, and potential implications. As an illustrative example, consider the operationalization of ``offensive'' across several offensive language datasets~\citep{warner2012detecting,rosenthal2021solid,nobata2016abusive}, where authors employ a keyword-based approach, selecting texts containing slurs or swears~\citep{waseem2017understanding}. Defining ``offensiveness'' based on keywords presents challenges, as it may overlook \emph{implicitly} offensive text~\citep{Wiegand_2021,waseem2017understanding} or language where the true offensive nature is obscured~\citep{Hada_2021}. Further, this operationalization ignores the context of the text, leading to potential mislabeling. For example, depending on the text's author and context, the use of an identity term can be either innocuous or pejorative~\citep{dixon2018measuring,davidson2019racial}.


\smallsec{Evaluate trade-offs for data collection}
Curators should thoroughly assess various data collection strategies and justify their chosen method. Trade-offs are inevitable, and while demonstrating robustness to different measurement types is ideal, it can be challenging, especially in resource-intensive dataset collection. At a minimum, transparent communication of decision-making processes facilitates the evaluation of operationalization and collection process validity. For example,~\citet{Zhao_2021} revealed a demographically imbalanced annotator pool, a limitation when crowdsourcing skin color and gender expression annotations on Amazon Mechanical Turk. Sharing insights into both successful and unsuccessful methods can prevent the duplication of null or negative results by future curators.

\section{Reliability}
\label{sec:reliability}

Measurement evaluation involves assessing two critical qualities: reliability and validity. In this section, we specifically delve into the concept of reliability, which concerns the consistency and dependability of measurement results. For diverse dataset collection, ensuring the reliability of the dataset is pivotal, forming the bedrock for the validity of the entire collection process---that is, how effectively the gathered dataset captures the essence of diversity.

To assess reliability in diverse datasets, we illuminate two evaluation methods drawn from the literature on measurement theory: \emph{inter-annotator agreement} and \emph{test-retest reliability}. We conceptualize dataset reliability as analogous to concerns about the quality and consistency of the collection process. Much like with an unreliable measurement, a dataset that yields inconsistent results when applied to the same case erodes trust in drawn conclusions, making the reliability of data collection essential~\citep{bandalos2018measurement}.




\subsection{Considerations} 

\smallsec{Lack of details on quality control} 
In our corpus analysis, a notable theme that emerged is the limited information available on quality control measures for datasets. Only 56.3\% of the datasets provided specifics about their quality control processes. This deficiency, as discussed in~\Cref{sec:operationalize}, is closely tied to the broader issue of inadequate details in the dataset collection process.

\smallsec{Erasure of labor} 
While datasets occasionally offer information regarding quality, the focus tends to be on human annotators rather than the data instances. Surprisingly, only 36.8\% of datasets include crucial details about annotation quality, such as annotator training processes, attention checks, compensation methods, and work rejection policies. This omission underscores the undervaluation of the labor invested in dataset creation. Crowdworkers are often seen as costs to be minimized, overlooking their significant contributions to datasets~\citep{williams2022exploited,shmueli2021beyond,miceli2022studying}. Similarly, the substantial effort required to ensure high-quality datasets is frequently ignored.

\subsection{Recommendations}
\label{sec:subsec:reliablityrec}

\smallsec{Inter-annotator agreement} 
An established method for assessing reliability, particularly in crowdsourcing, is through inter-annotator agreement. This method often entails multiple annotators labeling an instance, with the final label determined by a majority vote~\citep{davani2022dealing}. Another method to gauge inter-annotator reliability is by employing statistical measures of agreement. We find that some text datasets provide quantitative metrics~\citep{Sun_2021,Castro_2022,Cao_2020,Peskov_2019,Johnson_2018,Angelidis_2018,ide2008masc,Webster_2018,Zhong_2021}, such as Fleiss's $\kappa$~\citep{fleiss1971measuring} or Cohen's $\kappa$~\citep{cohen1960coefficient}, to quantify inter-annotator agreement. While consensus methods are employed in both text and image datasets, quantitative metrics for inter-annotator agreement are reported exclusively in text datasets. We recommend that image dataset curators also incorporate these statistical measures when evaluating crowdsourced labels.

The suitability of inter-annotator agreement is contingent on how diversity is defined and operationalized in the specific dataset. While measuring agreement enhances reliability, in certain cases, it might conflict with the diversity goal. For example, prior research has demonstrated that ``integrating dissenting voices into machine learning models''~\citep{gordon2022jury} can be beneficial, conferring ``robustness to adversarial attacks''~\citep{peterson2019human}. Furthermore, relying on aggregation methods, such as majority voting, may systematically erase certain groups' perspectives from datasets altogether~\citep{davani2022dealing,sachdeva2022assessing}. This tension between reliability and diversity is fundamentally linked to the chosen construct of diversity. If diversity involves a range of annotator demographics or perspectives, traditional reliability measures may not be appropriate. However, in tasks aiming for singular, congruous ``objective'' labels, incorporating inter-annotator agreement metrics can offer valuable insights into reliability.



\smallsec{Test-retest reliability} 
Another approach that dataset collectors can adopt is the test-retest method. In education, this method involves administering the same test twice over a period, with consistent results indicating reliability~\citep{guttman1945basis}. This principle is particularly relevant when assessing the reliability of collection methods like web scraping. For instance, curators can reapply the same methodology to recollect instances, validating whether the recollected dataset maintains the same diversity properties. Nonetheless, as emphasized by \citet{jacobs2021measurement}, a lack of reliability from these tests does not necessarily imply that the collection methodology inadequately captures diversity. Changes in the underlying data distribution over time~\citep{chen2023twigma,yao2022wild} can influence the results. For example, when evaluating linguistic diversity using data scraped from Reddit, major societal events, such as elections~\citep{waller2021quantifying}, can unexpectedly alter the distribution. Even in such cases, measuring test-retest reliability remains valuable for gaining insights into potential shifts in data distributions.

\section{Validity}
\label{sec:validity}

Conceptually, construct validity ensures that a measure \emph{aligns} with theoretical hypotheses. Adapting this definition, we explore the construct validity of diversity in ML datasets, aiming to determine whether the final dataset aligns with theoretical definitions. To assess validity in diverse datasets, in this section, we apply two commonly used subtypes of construct validity~\citep{campbell1959convergent} for evaluation: \emph{convergent validity} and \emph{discriminant validity}.



\subsection{Considerations}

\smallsec{Lack of robust validation} 
We observe a lack of robust validation for diversity claims made by dataset creators. This issue is partly rooted in the ambiguity surrounding the term. When the construct lacks a clear definition, it becomes challenging to empirically assess whether the collected dataset genuinely adheres to the specified standards. Even when validation is attempted, it may often center around incorrect constructs. While dataset authors may present metadata frequency or other summary statistics about their datasets, these metrics do not consistently align with the diversity dimensions described during the dataset's motivation

\smallsec{Overreliance on downstream evaluation} Another prevalent evaluation method involves benchmarking the downstream performance of a newly proposed model. This approach is observed in 48.5\% of datasets. However, it may assess the wrong construct by primarily focusing on model performance rather than the intrinsic characteristics of the dataset. Model performance, for instance, may increase due to the learning of ``shortcuts''~\citep{geirhos2020shortcut} rather than genuine increases in dataset diversity.

\subsection{Recommendations}

\smallsec{Convergent validity}
Convergent validity ensures correlated results for different measurements of the same construct~\citep{campbell1959convergent}. One approach is to compare a newly collected dataset to existing ones. Striking the right balance involves establishing the novelty of a new dataset while demonstrating its similarity to existing work~\citep{quinn2010analyze,jacobs2021measurement}. We provide recommendations for evaluating convergent validity.

\smallsecemph{Cross-dataset generalization} Commonly employed to evaluate ``dataset bias''~\citep{torralba2011unbiased}, cross-dataset generalization enables researchers to compare datasets. By utilizing existing datasets with similar structures (e.g., label taxonomy, modality) and constructs of diversity, collectors can train on their dataset and test on existing datasets or vice versa, comparing relevant metrics such as accuracy. Model performance can also be assessed against standard train-test splits from the same dataset. If the models perform similarly in both cross-dataset and same-dataset scenarios, it suggests that the datasets have similar distributions for the target variable, indicating correlated constructs of diversity. Prior work~\citep{khan2021one} adopted Fleiss's $\kappa$ to measure the consistency of predictions among models. However, a constraint of employing cross-dataset generalization is the necessity for congruent taxonomies (for the target variable) and comparable distributions across datasets.

\smallsecemph{Comparing existing diversity metrics} Dataset collectors can leverage established metrics for measuring data diversity~\citep{mitchell2022measuring}. For instance, \citet{friedman2022vendi} introduced the Vendi Score, drawing inspiration from ecology and quantum statistical mechanics, as a measure of diversity within image and text dataset categories. Curators can demonstrate how their collection process aligns with such recognized diversity metrics. Given that diversity metrics depend on the embedding space employed~\citep{friedman2022vendi}, datasets should be benchmarked across a multiplicity of spaces optimized for the definition of diversity selected by the dataset curators.

\smallsec{Discriminant validity}
Discriminant validity assesses whether measurements for theoretically unrelated constructs yield unrelated results~\citep{campbell1959convergent}. Consider the initial Visual Question Answer (VQA) dataset~\citep{antol2015vqa}, which aimed to collect diverse and interesting questions and answers, encompassing question types such as ``What is ...'', ``How many ...'', and ``Do you see a ...''. If diversity is defined by the types of questions asked, it should have no relation to other factors, such as gender distribution.\footnote{Note that comparing \emph{gender distribution} may not be reasonable for testing discriminant validity when the interest is in \emph{demographic diversity}, as these are related constructs.} 

Prior works~\citep{goyal2017making,agrawal2018don,Hudson_2019,Johnson_2017} identified language biases in how questions and answers are formulated in the VQA dataset. For instance, based on the dataset construction, a model predicting ``Yes'' whenever the question begins with ``Do you see a ...'' can achieve high accuracy without considering the image in question~\citep{goyal2017making}. This suggests potential low discriminant validity for the given measure, highlighting the importance of applying discriminant validity to mitigate construction biases during dataset creation.

\section{Case Study}
\label{sec:case_study}
To demonstrate the practical application of measurement theory in ML data collection, we examine the Segment Anything dataset (SA-1B)~\citep{kirillov2023segment} as a case study. We choose SA-1B due to its recent release and its alignment with large-scale datasets aimed at advancing foundation model research~\citep{bommasani2021opportunities}. Moreover, SA-1B benefits from clear documentation~\citep{gebru2021datasheets} provided by the authors. Through this case study, we investigate the conceptualization, operationalization, and evaluation of diversity, showcasing how our recommendations can enhance the dataset collection process.

SA-1B comprises ``11M diverse, high-resolution, licensed, and privacy protecting images and 1.1B high-quality segmentation masks''. According to the datasheet, the authors emphasize the dataset's enhanced geographical diversity compared to existing datasets, driven by the goal of fostering ``fairer and more equitable models''. The images are sourced from a third-party and captured by photographers. Notably, the masks are generated using a novel data engine detailed in the paper, and they are not semantically labeled.

\smallsec{Conceptualization} 
Although the datasheet mentions how diversity is defined, we can enhance this conceptualization by providing a more concrete and specific definition. First, the term ``geographic diversity'' is clarified to encompass a variety in both the country where the image is taken and the socioeconomic status of that country. Second, beyond geographic diversity, the dataset also defines diversity in terms of the variety in object appearance, including factors such as object size and complexity of object shape, as well as the number of objects per image. Finally, these definitions can be contextualized by referencing existing geodiverse datasets~\citep{rojas2022dollar,shankar2017no}. 

\smallsec{Operationalization}
Transitioning from conceptualization to operationalization, the different variables are implemented as outlined below. We highlight a strength of this work, which lies in the clear and well-defined indicators.
\begin{itemize}[leftmargin=2mm,noitemsep,topsep=-2pt]
    \item \textit{Country of origin}: Inferred from a caption describing the content in the image and a named-entity recognition model (NER)~\citep{peters2017semi} to identify location names.
    \item \textit{Socioeconomic status}: Used the World Bank's~\citep{WDI} income level categorization for the country.
    \item \textit{Object size}: Calculated image-relative mask size (i.e., square root of mask area divided by image area).
    \item \textit{Object complexity}: Calculated mask concavity (i.e., 1 - mask area divided by area of masks' convex hull).
    \item \textit{Number of objects}: Calculated as the count of masks.
\end{itemize}
A critical examination of these operationalizations brings forth two noteworthy considerations. First, as acknowledged by the authors, relying on an NER model for inferring the country of origin introduces the potential for errors due to, e.g., social bias or ambiguity (e.g., ``Georgia'' can refer to both a US state and a country). Second, opting to operationalize geolocation at the country level overlooks intra-national differences, potentially leading to the presentation of stereotypical representations of individuals within that country~\citep{naggita2023flickr}.

While the indicators are well-defined, there is room for improvement in the transparency surrounding the dataset collection process. The dataset, collected through a third party, reflects increasing opacity in collection processes (\Cref{subsec:operationalization_considerations}). Several important details about the collection process are omitted. For instance, the instructions given to photographers to enhance variation in object appearance and location are undisclosed. Further, while SA-1B does achieve diversity in objects, it is unclear whether this is a result of explicit instructions and deliberate sampling or simply a byproduct of scale.

\smallsec{Evaluation}
The authors substantiate their diversity claims by comparing object masks with those of similar datasets, such as COCO~\citep{lin2014microsoft} and OpenImages~\citep{kuznetsova2020open}. For example, they demonstrate alignment with respect to the distribution of object complexity. SA-1B distinguishes itself, however, by exhibiting more variety in mask sizes and the number of masks per image, showcasing convergent validity. 

A recommended enhancement involves providing validation for geographic diversity. Although the authors present figures illustrating the distribution of country of origin, emphasizing geodiversity is crucial as object appearances vary globally~\citep{de2019bias,shankar2017no,ramaswamy2023beyond}. To bolster the validation of this diversity aspect, the authors could compare distributions inferred from object labels or visualize object mask representations across different geographic regions.

\section{Discussion}
Finally, we dive into additional considerations regarding three key points: the inherent tensions between measurement and scale, the documentation burden imposed on dataset creators, and changing constructs over time.

\smallsec{Measurement and scale}
One potential tension within our proposed framework revolves around the interplay between diversity and scale. Currently, there is a prevalent belief among dataset curators that diversity will organically emerge as a consequence of dataset scale. For instance, \citet{sindagi2019pushing} argue that having ``such a large number of images'', in their dataset, JHU-Crowd, ``results in increased diversity in terms of count, background regions, scenarios, etc.'' Contrary to this notion, we posit that diversity, along with other constructs dataset curators aim to capture, is not an automatic byproduct of scale. Instead, it requires careful conceptualization, operationalization, and subsequent evaluation. This may impede scalability by introducing an additional need for curation and explicit control on the curator's side. Nevertheless, as previous studies have advocated~\citep{diaz2023scaling,hanna2020against}, challenging the concept of ``scale thinking'' can be advantageous, not only ethically but also for downstream performance. For example, \citet{Byrne_2021} emphasize the importance of ``carefully curated, annotated datasets that cover all the idiosyncrasies of a single task or transaction'' as a key factor in enhancing model performance, countering the prevailing notion solely focused on scale.

\smallsec{Documentation burden} 
In \Cref{subsec:operationalization_considerations}, we shed light on the \textit{documentation gap} in ML research. To tackle this challenge, we advocate for enhanced clarity and explicit communication throughout the dataset collection pipeline. However, we acknowledge the significant and often underestimated burden that documentation places on dataset creators---a task that is frequently undervalued within research communities and organizations~\citep{heger2022understanding}. This burden is further exacerbated by the necessity to address existing ``documentation debt''~\citep{bandy2021addressing}, referring to commonly used datasets with inadequate or absent documentation. Overcoming these challenges demands systemic changes in how data work is perceived within the ML community~\citep{sambasivan2021everyone}. We highlight ongoing efforts in this area, including the establishment of academic venues such as the Journal of Data-centric Machine Learning Research and NeurIPS Datasets and Benchmarks Track. These initiatives mark a crucial initial step toward tackling this pervasive problem.

\smallsec{Constructs changing over time}
Our framework does not account for potential shifts over time, such as discrepancies between training and testing data caused by temporal changes~\citep{yao2022wild,bergman2023representation}. In~\Cref{sec:subsec:reliablityrec}, we acknowledge this when discussing how the reliability of webscraped data can be impacted by changes to Internet trends or current events. Moreover, the very nature of the construct being measured may undergo transformations. For instance, the racial categories employed in the US Census have undergone significant changes over the years~\citep{Lai_Medina_2023}. Since these categories often serve as a taxonomy for race in ML datasets, it is evident how such changes can influence and date the underlying construct. Recognizing that construct definitions should be contextualized, datasets inevitably become tied to the temporal setting of their collection. Instead of striving for time-invariant datasets, we advocate for directing our efforts towards the development of algorithms capable of withstanding such distribution shifts~\citep{koh2021wilds,yao2022wild}.

\section{Conclusion}

We explored the application of measurement theory principles as a framework for enhancing ML datasets. Present data collection practices often treat value-laden constructs within datasets, like diversity, as implicit or self-explanatory. This approach gives rise to subsequent challenges in validating and replicating the assertions made by authors in their work. Significantly, the lack of a standardized framework in dataset creation exacerbates issues related to the reproducibility crisis in science. Our analysis discerns that the absence of clear definitions and quantification by dataset authors amounts to \emph{selective reporting}, hindering standardization. By leveraging measurement theory principles, we advocate for a more nuanced and precise approach to conceptualizing, operationalizing, and evaluating value-laden properties in dataset construction. This approach not only fosters transparency, reliability, and reproducibility in ML research but also contributes to addressing the pervasive issue of the reproducibility crisis in science.

\newpage
\section*{Impact Statement}
Our paper introduces strategies for enhancing ML dataset collection by leveraging measurement theory principles. By fostering a more deliberate and thoughtful approach among researchers, our work stands to positively influence the societal impact of ML dataset curation. Moreover, our recommendations extend beyond mere dataset collection, offering insights into improving transparency and reproducibility within ML research more broadly. 

\section*{Acknowledgements}
This work was funded by Sony Research. We would like to thank Will Held, Camille Harris, Omar Shaikh, and Chenglei Si for their helpful comments and suggestions.

\bibliography{egbib,dataset}
\bibliographystyle{icml2024}

\clearpage
\appendix
\onecolumn
\section{Methodology}
\label{sec:appendix_method}
We provide additional details on our methodology, including how we selected datasets to review, our coding procedure, and categorization details. 

\subsection{Literature Review}
\begin{figure}[h!]
    \centering
    \includegraphics[width=\linewidth]{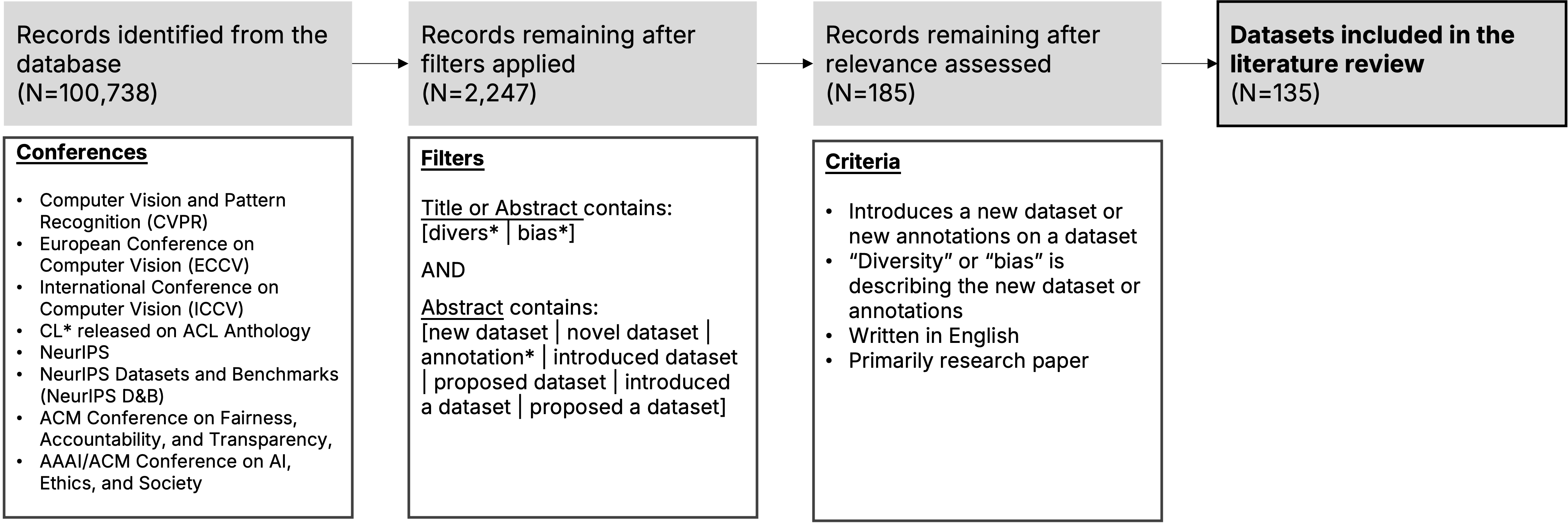}
    \caption{Overview of the search strategy used to identify diverse datasets for inclusion in our corpus.}
    \label{fig:methods}
\end{figure}

We surveyed a total of 135 image and text datasets using a defined process outlined in \Cref{fig:methods}.  Initially, we collected papers from key conferences and journals in computer vision (CVPR, ECCV, ICCV), natural language processing (*CL released on ACL Anthology), machine learning (NeurIPS, NeurIPS Datasets and Benchmarks Track), and fairness (FAccT, AIES) venues. Using regular expressions, we identified papers containing keywords related to ``diversity'' or ``bias'' in their titles or abstracts. Subsequently, we filtered these papers based on abstract content, focusing on those introducing new datasets or annotations. Manual screening ensured that the discussions on ``diversity'' or ``bias'' were pertinent to dataset creation, the articles were in English, and primarily research-focused. Finally, we randomly selected 135 datasets from this pool for review.

\definecolor{pastelgray}{rgb}{0.81, 0.81, 0.77}
\begin{table}[b!]
\scriptsize
    \centering
    \renewcommand{\arraystretch}{2}
    \begin{tabularx}{\textwidth}{@{}lX@{}}
    \toprule
    Dataset &  Diversity Definition \\ \midrule
    Ithaca365~\cite{diaz2022ithaca365} & ``[D]ata is repeatedly recorded along a 15 km route under diverse scene (urban, highway, rural, campus), weather
(snow, rain, sun), time (day/night), and traffic conditions
(pedestrians, cyclists and cars)''\\
    Vehicle Re-Identification for Aerial Image (VRAI)~\cite{Wang_2019} & ``The images are taken by
two moving UAVs in real urban scenarios, and the flight altitude ranges from 15m to 80m. It results in a large diversity of view-angles and pose variations, and so increases the
difficulty of the corresponding ReID task.''\\
GTA5 Crowdcounting~\cite{Wang_2019_Crowd}& ``Diverse Scenes. GCC dataset consists of 400 different
scenes, which includes multiple types of locations. For example, indoor scenes: convenience store, pub, etc. outdoor
scenes: mall, street, plaza, stadium and so on ... Diverse Environments. In order to construct the data
that are close to the wild, the images are captured at a random time in a day and under a random weather conditions.
In GTA5, we select seven types of weathers: clear, clouds,
rain, foggy, thunder, overcast and extra sunny.'' \\
Bilingual Text Separation (BTS)~\cite{xu2020diversity} & ``The
diversity of BTS can be described at three levels: (1) scene-level diversity: it covers common life scenes including
street signs, shop signs, plaques, attractions, book covers,
banners, and couplets; (2) image-level diversity: appearances and geometric variances caused by camera-captured
settings and background distractions such as perspective,
illumination, resolution, partly blocking, blur and so on,
in total including 14,250 fine-annotated text images; (3)
character-level diversity: variances of character categories,
up to 3,985 classes including Chinese characters, English
letters, digits, common punctuation with varied fonts and
sizes.''\\

    \bottomrule
    \end{tabularx}
    \caption{Example quotes illustrating concrete definitions of diversity from selected surveyed datasets.}
    \label{tab:sample}
\end{table}

\subsection{Coding Procedure}
Coding was performed by DZ, JTAA, and OP. We initially coded a set of three datasets independently. After coding these initial datasets, we collaboratively discussed any discrepancies in our codes and refined the codebook definitions. Subsequently, we divided the remaining 132 datasets among the three of us for individual coding. To identify overarching themes, each of us individually generated themes, which we later grouped into larger categories during synchronous discussions.

\subsection{Categorization}
\smallsec{Definition of standalone dataset}
A paper was considered to present a standalone dataset if its primary contributions were the introduction of a novel dataset, benchmark, or set of annotations. For example, if a paper introduces a new model, task, or algorithm, it would \textit{not} be classified as presenting a standalone dataset.

\smallsec{Criteria for concrete definition}
We applied the following criteria to determine whether a concrete definition of diversity was provided in a dataset paper. We checked whether the dataset authors explicitly stated a consistent and complete definition of diversity. For example, if a dataset was described as ``diverse'' in multiple ways throughout the paper, it would not be considered to have a concrete definition. Additionally, we ensured that the dataset authors listed out the specific aspects that were more diverse (e.g., different illumination conditions, sources of news). Examples of concrete diversity definitions identified are provided in \Cref{tab:sample}.

\section{Additional Dataset Collection Details}
In~\Cref{sec:operationalize}, we categorize dataset collection methodologies into five main types. Further details on this taxonomy are provided in \Cref{tab:collection_method}, with a list of datasets in our corpus categorized accordingly in \Cref{tab:collection_list}.
\begin{table}[b!]
    \centering
    \renewcommand{\arraystretch}{2}
    \scriptsize
    \begin{tabularx}{\textwidth}{@{} l X X @{}}
    \toprule
    Method &  Definition & Example \\
    \midrule
    Webscraping ($N=47$)& Instances are sourced and downloaded from content that is posted to the Internet, typically for purposes other than being used for creating a machine learning dataset &         
    \begin{minipage}[t]{\linewidth}
        \begin{itemize}[leftmargin=3mm,noitemsep,topsep=0pt]
            \item \textbf{Text}: Mega-COV~\citep{Abdul_Mageed_2021} consists of multilingual tweets sourced using Twitter's streaming API.  
            \item \textbf{Image}: \citet{Idrees_2018} combine three web sources, ``Flickr, Web Search, and the Hajj footage,'' to create UCF-QNRF, an image dataset used for crowd counting.
        \end{itemize}
    \end{minipage}
    \\
    Crowdsourcing ($N=9$)& Instances are sourced from a group of people that are not directly members of the research team for the purposes of creating a machine learning dataset &     
    \begin{minipage}[t]{\linewidth}
        \begin{itemize}[leftmargin=3mm,noitemsep,topsep=0pt]
            \item \textbf{Text}: To create dialogue data for MultiDoGO, \citet{Peskov_2019} ask Amazon Mechanical Turk workers to engage with an ``agent'' in an Wizard-of-Oz approach. 
            \item \textbf{Image}: EPIC-KITCHENS~\citep{Damen_2018} contains video footage collected by 32 participants across four different countries. To note, participants contribute this footage voluntarily and without compensation. 
        \end{itemize}
    \end{minipage}
    \\
    Direct Collection ($N=20$)& Instances are created or collected by members of the research team in the real-world &   
    \begin{minipage}[t]{\linewidth}
        \begin{itemize}[leftmargin=3mm,noitemsep,topsep=0pt]
            \item \textbf{Text}: While there are no text datasets from our corpus in which the instances are directly collected, the GICoref~\citep{Cao_2020} is an example where annotations are manually generated (i.e., directly collected) by the paper authors. 
            \item \textbf{Image}: The Waymo Open Dataset~\citep{Sun_2020} is collected using specialized equipment and vehicles across three different locations: Phoenix, Mountain View, and San Francisco. 
        \end{itemize}
    \end{minipage}
    \\
    Derivative ($N=47$)& Instances are sourced from an already existing dataset &         
    \begin{minipage}[t]{\linewidth} \begin{itemize}[leftmargin=3mm,noitemsep,topsep=0pt]
            \item \textbf{Text}: \citet{Ni_2019} combine review datasets from Amazon Clothing and Yelp Reviews dataset to generate a novel dataset for recommendation justification.
            \item \textbf{Image}: The Localized Narratives dataset~\citep{Pont_Tuset_2020} provides new annotations on images sourced from COCO~\citep{lin2014microsoft}, Flickr30k~\citep{plummer2015flickr30k}, ADE20K~\citep{zhou2017scene}, and Open Images~\citep{kuznetsova2020open}. 
        \end{itemize}
        \end{minipage}\\
    Synthetic Generation ($N=13$)& Instances are artificially manufactured or procedurally generated rather than capturing real-world events &     \begin{minipage}[t]{\linewidth}
        \begin{itemize}[leftmargin=3mm,noitemsep,topsep=0pt]
            \item \textbf{Text}: SynthBio~\citep{yuan2021synthbio} is generated using language models which involve creating attribute lists (e.g., notabilities, nationality, birth date) and synthesizing biographies for people who possess these attributes. 
            \item \textbf{Image}: \citet{karazija2021clevrtex} create the image scenes in ClevrTex using a selection of photo-mapped materials, objects, and backgrounds that are procedurally generated with the Blender API.  
        \end{itemize}
    \end{minipage}
    \\
    \bottomrule
    \end{tabularx}
    \caption{Taxonomy of the commonly used dataset collection methodologies identified through our literature review. We provide a definition of the collection methodology, the number of datasets ($N$) in our corpus that use this methodology, as well as an example of an image and text dataset from our corpus.}
    \label{tab:collection_method}
\end{table}

\begin{table}[]
\scriptsize
    \centering
    \renewcommand{\arraystretch}{2}
    \begin{tabularx}{\textwidth}{@{}lX@{}}
    \toprule
        
         Method &  Dataset\\ \midrule
         Webscraping &  VideoCoAtt~\cite{Fan_2018}, VCR~\cite{Zellers_2019}, StockEmotion~\cite{wei2020learning}, FineDiving~\cite{xu2022finediving}, iNat2021~\cite{Van_Horn_2021}, Natural World Tasks (NeWT)~\cite{Van_Horn_2021}, FEVEROUS~\cite{aly2021feverous}, Semantic Scholar Visual Layoutenhanced Scientific Text Understanding Evaluation
        (S2-VLUE)~\cite{Shen_2022}, Mega-COV~\cite{Abdul_Mageed_2021}, Bajer~\cite{Zeinert_2021}, GICOREF~\cite{Cao_2020}, StackEx~\cite{Yuan_2020}, \citet{Gillani_2019}, DuConv~\cite{Wu_2019}, \citet{Durmus_2019}, Broad Twitter Corpus (BTC)~\cite{derczynski2016broad}, VPS~\cite{Miao_2021}, JHU-Crowd~\cite{sindagi2019pushing}, PASS~\cite{asano2021pass}, LiRo~\cite{dumitrescu2021liro}, IconQA~\cite{lu2021iconqa}, \citet{Pratapa_2022}, FairytaleQA~\cite{Xu_2022}, \citet{celis2016fair}, Ruddit~\cite{Hada_2021}, CMU-MOSEAS (CMU Multimodal Opinion Sentiment, Emotions and Attributes)~\cite{Bagher_Zadeh_2020}, Video-and-Language Event Prediction (VLEP)~\cite{Lei_2020}, Big BiRD: A Large, Fine-Grained, Bigram Relatedness Dataset for Examining Semantic Composition~\cite{asaadi2019big}, Reddit TIFU~\cite{Li_2019}, MAD~\cite{Soldan_2022}, \citet{Fu_2021}, Mapillary Traffic Sign Dataset (MTSD)~\cite{Ertler_2020}, TrackingNet~\cite{M_ller_2018}, UCF-QNRF~\cite{Idrees_2018}, SketchyScene~\cite{Zou_2018}, Video gAze CommunicATION (VACATION)~\cite{Fan_2019}, Query-based Video Highlights (QVHighlights)~\cite{lei2021detecting}, \citet{rahman2021information}, Chinese LOng Text understanding andgeneration (LOT)~\cite{Guan_2022}, WikiEvolve~\cite{De_Kock_2022}, \citet{Sugawara_2022}, Multi-lingual retrieval Typologically Diverse (Mr. TyDi)~\cite{Zhang_2021}, WIKIBIAS~\cite{Zhong_2021}, Gender Shades~\cite{buolamwini2018gender}, iEAT~\cite{Steed_2021}, GAP Coreference~\cite{Webster_2018}, Bias in Bios~\cite{de2019bias} \\
         Crowdsourcing & EPIC-KITCHENS~\cite{Damen_2018}, ObjectNet~\cite{barbu2019objectnet}, MultiDoGO~\cite{Peskov_2019}, SCT v1.5~\cite{Sharma_2018}, Charades~\cite{Sigurdsson_2016}, \citet{Wiegand_2021}, TicketTalk~\cite{Byrne_2021}, Taskmaster-1~\cite{Byrne_2019}, Camp Site Negotiation (CaSiNo)~\cite{Chawla_2021} \\
         Direct Collection & ETH3D~\cite{Schops_2017}, Cityscapes~\cite{Cordts_2016}, SUNRGB-D~\cite{Song_2015}, RID~\cite{Wan_2018}, \citet{Wang_2019}, Waymo Open Dataset~\cite{Sun_2020}, \citet{Zhang_2020}, HOI4D~\cite{Liu_2022}, Person30k~\cite{Bai_2021}, VisDrone-DET 2018~\cite{Cao_2021}, Vehicle Re-Identification for Aerial Image (VRAI)~\cite{Wang_2019}, Drive\&Act~\cite{Martin_2019}, Wilddash2 (WD2)~\cite{Zendel_2022}, VTUAV~\cite{Zhang_2022}, Nutrition5k~\cite{Thames_2021}, Habitat-Matterprot 3D (HM3D)~\cite{Yadav_2023}, Ithaca365~\cite{diaz2022ithaca365}, Multimodal Audio-Visual Detection (MAVD)~\cite{Valverde_2021}, FreiHAND~\cite{Zimmermann_2019}, Vision-based Fallen Person (VFP290K)~\cite{an2021vfp290k} \\
         Derivative & SUNRGB-D~\cite{Song_2015}, ADE-Affordance~\cite{Chuang_2018}, OK-VQA~\cite{Marino_2019}, GQA~\cite{Hudson_2019}, VCR~\cite{Zellers_2019}, DAVSOD~\cite{Fan_2019}, CoSOD3k~\cite{Fan_2020}, KeypointNet~\cite{you2020keypointnet}, ArtEmis v2.0~\cite{Mohamed_2022}, VidSitu~\cite{Sadhu_2021}, Tracking Any Object (TAO)~\cite{Dave_2020}, HowToVQA69M~\cite{Yang_2021}, iVQA~\cite{Yang_2021}, CropHarvest~\cite{tseng2021cropharvest}, FIBER~\cite{Castro_2022}, DocRED~\cite{Huang_2022}, \citet{Sun_2021}, YASO~\cite{Orbach_2021}, Mickey Corpus~\cite{Lin_2021}, X-CSQA~\cite{Lin_2021}, ConvoSumm~\cite{Fabbri_2021}, Room-Across-Room (RxR)~\cite{Ku_2020}, \citet{Ni_2019}, \citet{Johnson_2018}, OpoSum~\cite{Angelidis_2018}, Manually Annotated Sub-Corpus (MASC)~\cite{ide2008masc}, WIDERFACE-DEMO~\cite{Yang_2022}, TVR~\cite{Lei_2020}, PASS~\cite{asano2021pass}, LiRo~\cite{dumitrescu2021liro}, AP-10K~\cite{yu2021ap}, \citet{Drawzeski_2021}, MightyMorph~\cite{Goldman_2021}, Amazon Customer Reviews~\cite{O_Neill_2021}, mTVR~\cite{Lei_2021}, ArSarcasm~\cite{farha2020arabic}, R4C~\cite{Inoue_2020}, \citet{Lepp_2020}, \citet{Schulz_2019}, MAD~\cite{Soldan_2022}, VIdeo Panoptic Segmentation in the Wild (VIPSeg)~\cite{Miao_2022}, TrackingNet~\cite{M_ller_2018}, Localized Narratives~\cite{Pont_Tuset_2020}, \citet{Zhao_2021}, \citet{hemani2021ails}, \citet{Culkin_2021}, Racial Faces in the Wild (RFW)~\cite{Wang_2019} \\
         Synthetic Generation & CLEVR~\cite{Johnson_2017}, Synthia~\cite{Ros_2016}, GTA5 Crowd Counting~\cite{Wang_2019}, FakeAVCeleb~\cite{khalid2021fakeavceleb}, MotSynth~\cite{fabbri2021motsynth}, ClevrTex~\cite{karazija2021clevrtex}, SketchyScene~\cite{Zou_2018}, 3D Furnished Rooms with layOuts and semaNTics (3D-FRONT)~\cite{Fu_2021}, SynthBio~\cite{yuan2021synthbio}, Chinese LOng Text understanding andgeneration (LOT)~\cite{Guan_2022}, \citet{Culkin_2021}, Winobias~\cite{Zhao_2018}, Winogender~\cite{Rudinger_2018}\\
         \bottomrule
    \end{tabularx}
    \caption{Collection methodologies employed in the datasets encompass a range of approaches: \emph{webscraping}, involving the retrieval of instances from the Internet; \emph{crowdsourcing}, which entails soliciting contributions from individuals not affiliated with the research team; \emph{direct collection}, where data instances are gathered by members of the research team themselves; \emph{derivative}, which leverages one or multiple existing datasets; and \emph{synthetic generation}, wherein instances are artificially generated rather than derived from real-world events. Note that each dataset may utilize multiple collection methodologies.}
    \label{tab:collection_list}
\end{table}

\section{Dataset Details}
\label{sec:appendix_dataset}
Here, we offer additional insights into the 135 image and text datasets coded during our literature review. \Cref{tab:diversity} provides a taxonomy for different types of diversity we observed and \Cref{tab:diversity_list} outlines the diversity type for each dataset. Furthermore, \Cref{tab:papers,tab:papers2,tab:papers3} summarize our analysis across all datasets. 

\begin{table}[b!]
    \centering
    \renewcommand{\arraystretch}{2}
    \scriptsize
    \begin{tabularx}{\textwidth}{@{} l X X @{}}
    \toprule
        Diversity & Definition & Example\\
    \midrule
       Composition (N=58)  & Variety in what a dataset instance contains, such as linguistic properties (e.g., language, vocabulary used), scene or background, objects, viewpoint, image properties (e.g., resolution, focal length), or pose
        & 
        \begin{minipage}[t]{\linewidth}
        \begin{itemize}[leftmargin=3mm,noitemsep,topsep=0pt]
            \item \textbf{Text}: X-CSQA~\citep{Lin_2021} is collected to extend question-answering text evaluation beyond just English into multiple other languages       
            \item \textbf{Image}: Nutrition5k~\citep{Thames_2021} contains  a ``wide variety of ingredients, portion sizes, and dish complexities.''
        \end{itemize}
        \end{minipage}
    \\
    Source (N=26) & Variety in where data instances are collected from, such as web source online or geographic origin & 
    \begin{minipage}[t]{\linewidth}
    \begin{itemize}[leftmargin=3mm,noitemsep,topsep=0pt]
        \item \textbf{Text}: ConvoSumm~\citep{Fabbri_2021} draws from multiple online sources (New York Times comments, Reddit, StackExchange, and email threads)
        \item \textbf{Image}: \citet{tseng2021cropharvest} sample their crop dataset across several countries
    \end{itemize}
    \end{minipage}
    \\ 
    Domain (N=18) & Variety in the ``topic area'' of the data instances, such as what disciplines the text is drawn from or what artistic style is represented in the image & 
    \begin{minipage}[t]{\linewidth}
    \begin{itemize}[leftmargin=3mm,noitemsep,topsep=0pt]
        \item \textbf{Text}: S2-VLUE~\citep{Shen_2022} consists of scientific papers from 19 different academic disciplines  
        \item \textbf{Image}: TVR~\citep{Lei_2020_TVR} contains videos from six TV shows across diverse genres
    \end{itemize}
    \end{minipage}
    \\ 
     Subject (N=16) & Representation of human subjects in the dataset, such as by protected attributes (e.g., gender, race, age), physical characteristics (e.g., skin tone, weight, height), nationality, socioeconomic status, and language  & 
     \begin{minipage}[t]{\linewidth}
     \begin{itemize}[leftmargin=3mm,noitemsep,topsep=0pt]
        \item \textbf{Text}: SynthBio~\citep{yuan2021synthbio} is a synthetically generated evaluation set for WikiBio~\citep{lebret2016neural} which is balanced with respect to the gender and nationality of biography subjects
        \item \textbf{Image}:  \citet{an2021vfp290k} recruit participants with different ``skin color and age'' as data subjects
    \end{itemize}
    \end{minipage}
     \\
     Annotator (N=2) & Representation of annotator backgrounds, such as demographic background, domain expertise, or political affiliation & 
    \begin{minipage}[t]{\linewidth}
     \begin{itemize}[leftmargin=3mm,noitemsep,topsep=0pt]
         \item \textbf{Text}: \citet{Zeinert_2021} recruit annotators with diversity ``in gender, age, occupation / background (linguistic and ethnographic knowledge), region (spoken dialects)'' to label misogyny online.
         \item \textbf{Image}: N/A
     \end{itemize}
    \end{minipage}
    \\
    \bottomrule
    \end{tabularx}
    \caption{Taxonomy of the definitions of diversity identified through our literature review. We provide a definition of diversity, the number of datasets (N) in our corpus that use the respective definition of diversity, as well as an example of an image and text dataset from our corpus. We do not find any image datasets in our corpus that seek to use a more diverse annotator pool.}
    \label{tab:diversity}
\end{table}

\begin{table}[t!]
\scriptsize
    \centering
    \renewcommand{\arraystretch}{2}
    \begin{tabularx}{\textwidth}{@{}lX@{}}
    \toprule
    Category & Dataset\\ \midrule
     Composition Diversity &  ETH3D~\cite{Schops_2017}, Cityscapes~\cite{Cordts_2016}, Synthia~\cite{Ros_2016}, ADE-Affordance~\cite{Chuang_2018}, RID~\cite{Wan_2018}, VideoCoAtt~\cite{Fan_2018}, OK-VQA~\cite{Marino_2019}, \citet{Wang_2019}, GTA5 Crowd Counting~\cite{Wang_2019}, DAVSOD~\cite{Fan_2019}, CoSOD3k~\cite{Fan_2020}, \citet{Zhang_2020}, HOI4D~\cite{Liu_2022}, Person30k~\cite{Bai_2021}, Tracking Any Object (TAO)~\cite{Dave_2020}, VisDrone-DET 2018~\cite{Cao_2021}, Vehicle Re-Identification for Aerial Image (VRAI)~\cite{Wang_2019}, ObjectNet~\cite{barbu2019objectnet}, Mickey Corpus~\cite{Lin_2021}, X-CSQA~\cite{Lin_2021}, Room-Across-Room (RxR)~\cite{Ku_2020}, Broad Twitter Corpus (BTC)~\cite{derczynski2016broad}, Wilddash2 (WD2)~\cite{Zendel_2022}, DAD-3DHeads~\cite{Martyniuk_2022}, Nutrition5k~\cite{Thames_2021}, VPS~\cite{Miao_2021}, Charades~\cite{Sigurdsson_2016}, Trans10K~\cite{Xie_2020}, \citet{Silberman_2012}, Waymo Open Motion Dataset~\cite{ettinger2021large}, MotSynth~\cite{fabbri2021motsynth}, JHU-Crowd~\cite{sindagi2019pushing}, Habitat-Matterprot 3D (HM3D)~\cite{Yadav_2023}, AP-10K~\cite{yu2021ap}, ClevrTex~\cite{karazija2021clevrtex}, \citet{Drawzeski_2021}, MightyMorph~\cite{Goldman_2021}, CMU-MOSEAS (CMU Multimodal Opinion Sentiment, Emotions and Attributes)~\cite{Bagher_Zadeh_2020}, Taskmaster-1~\cite{Byrne_2019}, MAD~\cite{Soldan_2022}, Bilingual Text Segmentation (BTS)~\cite{Xu_2022}, VIdeo Panoptic Segmentation in the Wild (VIPSeg)~\cite{Miao_2022}, Ithaca365~\cite{diaz2022ithaca365}, \citet{Fu_2021}, Multimodal Audio-Visual Detection (MAVD)~\cite{Valverde_2021}, Mapillary Traffic Sign Dataset (MTSD)~\cite{Ertler_2020}, TrackingNet~\cite{M_ller_2018}, Localized Narratives~\cite{Pont_Tuset_2020}, UCF-QNRF~\cite{Idrees_2018}, Cars Overhead with Context (COWC)~\cite{Mundhenk_2016}, 3D Furnished Rooms with layOuts and semaNTics (3D-FRONT)~\cite{Fu_2021}, FreiHAND~\cite{Zimmermann_2019}, Video gAze CommunicATION (VACATION)~\cite{Fan_2019}, Query-based Video Highlights (QVHighlights)~\cite{lei2021detecting}, Vision-based Fallen Person (VFP290K)~\cite{an2021vfp290k}, \citet{Culkin_2021}, Camp Site Negotiation (CaSiNo)~\cite{Chawla_2021}, Multi-lingual retrieval Typologically Diverse (Mr. TyDi)~\cite{Zhang_2021}\\
     Source Diversity & VideoCoAtt~\cite{Fan_2018}, Waymo Open Dataset~\cite{Sun_2020}, iNat2021~\cite{Van_Horn_2021}, EPIC-KITCHENS~\cite{Damen_2018}, Drive\&Act~\cite{Martin_2019}, CropHarvest~\cite{tseng2021cropharvest}, Semantic Scholar Visual Layoutenhanced Scientific Text Understanding Evaluation
    (S2-VLUE)~\cite{Shen_2022}, YASO~\cite{Orbach_2021}, Mega-COV~\cite{Abdul_Mageed_2021}, ConvoSumm~\cite{Fabbri_2021}, MultiDoGO~\cite{Peskov_2019}, \citet{Durmus_2019}, Broad Twitter Corpus (BTC)~\cite{derczynski2016broad}, Manually Annotated Sub-Corpus (MASC)~\cite{ide2008masc}, Wilddash2 (WD2)~\cite{Zendel_2022}, VTUAV~\cite{Zhang_2022}, The Dollar Street Dataset~\cite{rojas2022dollar}, PASS~\cite{asano2021pass}, Ithaca365~\cite{diaz2022ithaca365}, Multimodal Audio-Visual Detection (MAVD)~\cite{Valverde_2021}, Mapillary Traffic Sign Dataset (MTSD)~\cite{Ertler_2020}, Cars Overhead with Context (COWC)~\cite{Mundhenk_2016}, Video gAze CommunicATION (VACATION)~\cite{Fan_2019}, Query-based Video Highlights (QVHighlights)~\cite{lei2021detecting}, \citet{rahman2021information}, \citet{Sugawara_2022} \\
        Domain Diversity & OK-VQA~\cite{Marino_2019}, StockEmotion~\cite{wei2020learning}, FineDiving~\cite{xu2022finediving}, VidSitu~\cite{Sadhu_2021}, Tracking Any Object (TAO)~\cite{Dave_2020}, EPIC-KITCHENS~\cite{Damen_2018}, Drive\&Act~\cite{Martin_2019}, OpoSum~\cite{Angelidis_2018}, VPS~\cite{Miao_2021}, TVR~\cite{Lei_2020}, Waymo Open Motion Dataset~\cite{ettinger2021large}, GQA~\cite{Hudson_2019}, HowToVQA69M~\cite{Yang_2021}, IconQA~\cite{lu2021iconqa}, Video-and-Language Event Prediction (VLEP)~\cite{Lei_2020}, MAD~\cite{Soldan_2022}, \citet{Fu_2021}, Localized Narratives~\cite{Pont_Tuset_2020} \\
     Subject Diversity & \citet{Martin_2019, khalid2021fakeavceleb, Cao_2020, Yang_2022, rojas2022dollar, fabbri2021motsynth, Byrne_2021, Zimmermann_2019, an2021vfp290k, yuan2021synthbio, Chawla_2021, Wang_2019, buolamwini2018gender, Zhao_2018, Rudinger_2018, Webster_2018} \\
     Annotator Diversity & Bajer~\cite{Zeinert_2021}, mTVR~\cite{Lei_2021}\\ 
     Reduce Dataset Bias & CLEVR~\cite{Johnson_2017}, VCR~\cite{Zellers_2019}, ArtEmis v2.0~\cite{Mohamed_2022}, iVQA~\cite{Yang_2021}, FEVEROUS~\cite{aly2021feverous}, FIBER~\cite{Castro_2022}, DocRED~\cite{Huang_2022}, SCT v1.5~\cite{Sharma_2018}, Salient Objects in Clutter (SOC)~\cite{Fan_2018}, \citet{Wiegand_2021}, ArSarcasm~\cite{farha2020arabic}, R4C~\cite{Inoue_2020}, Big BiRD: A Large, Fine-Grained, Bigram Relatedness Dataset for Examining Semantic Composition~\cite{asaadi2019big}, Reddit TIFU~\cite{Li_2019}, \citet{hemani2021ails}, Chinese LOng Text understanding andgeneration (LOT)~\cite{Guan_2022}, WikiEvolve~\cite{De_Kock_2022}, WIKIBIAS~\cite{Zhong_2021} \\ 
     Promote Diversity (or Fairness) in Downstream Applications & Natural World Tasks (NeWT)~\cite{Van_Horn_2021}, \citet{Sun_2021}, StackEx~\cite{Yuan_2020}, \citet{Gillani_2019}, DuConv~\cite{Wu_2019}, \citet{Ni_2019}, \citet{Johnson_2018}, LiRo~\cite{dumitrescu2021liro}, FairytaleQA~\cite{Xu_2022}, \citet{celis2016fair}, \citet{Lepp_2020}, \citet{Zhao_2021}, iEAT~\cite{Steed_2021}, Bias in Bios~\cite{de2019bias} \\
     Not specified & SUNRGB-D~\cite{Song_2015}, KeypointNet~\cite{you2020keypointnet}, \citet{Pratapa_2022}, \citet{Schulz_2019}\\
        \bottomrule
    \end{tabularx}
    \caption{Surveyed datasets cover various categories of diversity, alongside additional identified factors such as \emph{reducing dataset bias} and \emph{promoting diversity (or fairness) in downstream applications}. Note that each dataset may encompass multiple categories of diversity.}
    \label{tab:diversity_list}
\end{table}

\begin{table}[]
    \scriptsize
    \centering
    \begin{tabular}{p{2.5in}p{0.3in}p{0.3in}p{0.3in}p{0.3in}p{0.3in}p{0.3in}}
    \rotatebox{0}{Dataset} & \rotatebox{45}{Standalone} & \rotatebox{45}{Concret defn.} & \rotatebox{45}{Justification} & \rotatebox{45}{Collection trade-offs} & \rotatebox{45}{Quality info} & \rotatebox{45}{Annot. info} \\
    \midrule
    3D Furnished Rooms with layOuts and semaNTics (3D-FRONT)~\citep{Fu_2021} &  &  &  &  &  & NA\\
    ADE-Affordance~\citep{Chuang_2018} &  &  & \checkmark &  &  & \\
    AP-10K~\citep{yu2021ap} & \checkmark & \checkmark & \checkmark &  & \checkmark & \checkmark\\
    Amazon Customer Reviews~\citep{O_Neill_2021} & \checkmark & NA & NA & \checkmark & \checkmark & \checkmark\\
    ArSarcasm~\citep{farha2020arabic} &  & NA & NA &  & \checkmark & \checkmark\\
    ArtEmis v2.0~\citep{Mohamed_2022} & \checkmark & NA & NA & \checkmark & \checkmark & \\
    Bajer~\citep{Zeinert_2021} & \checkmark & \checkmark & \checkmark & \checkmark & \checkmark & \checkmark\\
    Bias in Bios~\citep{de2019bias} &  & NA & NA & \checkmark &  & NA\\
    Big BiRD: A Large, Fine-Grained, Bigram Relatedness Dataset for Examining Semantic Composition~\citep{asaadi2019big} & \checkmark &  &  &  & \checkmark & \\
    Bilingual Text Segmentation (BTS)~\citep{Xu_2022} &  & \checkmark & \checkmark &  & \checkmark & \\
    Broad Twitter Corpus (BTC)~\citep{derczynski2016broad} & \checkmark & \checkmark & \checkmark & \checkmark & \checkmark & \checkmark\\
    CLEVR~\citep{Johnson_2017} & \checkmark & NA & NA &  &  & \\
    CMU-MOSEAS (CMU Multimodal Opinion Sentiment, Emotions and Attributes)~\citep{Bagher_Zadeh_2020} & \checkmark & \checkmark &  &  & \checkmark & \checkmark\\
    Camp Site Negotiation (CaSiNo)~\citep{Chawla_2021} &  &  &  & \checkmark &  & \\
    Cars Overhead with Context (COWC)~\citep{Mundhenk_2016} &  &  &  &  &  & \\
    Charades~\citep{Sigurdsson_2016} & \checkmark &  &  &  &  & \\
    Chinese LOng Text understanding andgeneration (LOT)~\citep{Guan_2022} &  & \checkmark & \checkmark &  & \checkmark & \\
    Cityscapes~\citep{Cordts_2016} & \checkmark & \checkmark & \checkmark & \checkmark & \checkmark & \checkmark\\
    ClevrTex~\citep{karazija2021clevrtex} & \checkmark & \checkmark & \checkmark &  &  & NA\\
    CoSOD3k~\citep{Fan_2020} & \checkmark & \checkmark &  &  & \checkmark & \checkmark\\
    ConvoSumm~\citep{Fabbri_2021} &  &  &  &  & \checkmark & \checkmark\\
    CropHarvest~\citep{tseng2021cropharvest} & \checkmark &  & \checkmark & \checkmark &  & NA\\
    DAD-3DHeads~\citep{Martyniuk_2022} &  & \checkmark &  &  & \checkmark & \\
    DAVSOD~\citep{Fan_2019} &  & \checkmark &  &  &  & \checkmark\\
    DocRED~\citep{Huang_2022} &  & NA & NA &  & \checkmark & \checkmark\\
    Drive\&Act~\citep{Martin_2019} & \checkmark &  & \checkmark &  &  & NA\\
    DuConv~\citep{Wu_2019} &  & NA & NA &  &  & NA\\
    EPIC-KITCHENS~\citep{Damen_2018} & \checkmark & \checkmark &  & \checkmark & \checkmark & \\
    ETH3D~\citep{Schops_2017} &  & \checkmark &  &  & \checkmark & NA\\
    FEVEROUS~\citep{aly2021feverous} &  & NA & NA & \checkmark & \checkmark & \checkmark\\
    FIBER~\citep{Castro_2022} &  & NA & NA & \checkmark & \checkmark & \checkmark\\
    FairytaleQA~\citep{Xu_2022_Fairy} & \checkmark &  &  &  & \checkmark & \checkmark\\
    FakeAVCeleb~\citep{khalid2021fakeavceleb} & \checkmark & \checkmark & \checkmark &  & \checkmark & NA\\
    FineDiving~\citep{xu2022finediving} &  &  & \checkmark &  &  & \\
    FreiHAND~\citep{Zimmermann_2019} &  & NA & NA &  &  & \\
    GAP Coreference~\citep{Webster_2018} &  & NA & NA &  & \checkmark & \\
    GICOREF~\citep{Cao_2020} &  & NA & NA &  & \checkmark & \\
    GQA~\citep{Hudson_2019} &  & \checkmark & \checkmark &  &  & \\
    GTA5 Crowd Counting~\citep{Wang_2019_Crowd} &  & \checkmark & \checkmark & \checkmark &  & NA\\
    Gender Shades~\citep{buolamwini2018gender} &  & NA & NA & \checkmark & \checkmark & \checkmark\\
    HOI4D~\citep{Liu_2022} & \checkmark & \checkmark & \checkmark &  &  & \\
    Habitat-Matterprot 3D (HM3D)~\citep{Yadav_2023} & \checkmark &  &  &  & \checkmark & \\
    HowToVQA69M~\citep{Yang_2021} &  &  &  &  & \checkmark & NA\\
    IconQA~\citep{lu2021iconqa} &  &  & \checkmark &  & \checkmark & \\
    Ithaca365~\citep{diaz2022ithaca365} & \checkmark & \checkmark & \checkmark &  &  & \\
    JHU-Crowd~\citep{sindagi2019pushing} &  & \checkmark &  &  &  & NA\\
    KeypointNet~\citep{you2020keypointnet} &  &  &  &  &  & \\
    \bottomrule
    \end{tabular}
    \caption{(Table continues in~\Cref{tab:papers2,tab:papers3}). Summary of our analysis across all 135 image and text datasets. Column labels: \textit{standalone}, indicating whether the dataset paper exclusively presents a dataset without introducing novel models, tasks, or algorithms; \textit{concrete definition}, indicating whether the dataset paper includes a clear definition of diversity (\emph{NA} for datasets focusing solely on bias, not diversity); \textit{justification}, denoting whether the dataset paper discusses the rationale behind incorporating diversity into the dataset (\emph{NA} for datasets focusing solely on bias, not diversity); \textit{collection trade-offs}, indicating whether the dataset paper discusses drawbacks of the chosen collection methodology or provides considerations for alternative methodologies; \textit{quality information}, signifying whether the dataset paper includes details on how dataset quality was verified (e.g., manual inspection by creators, inter-annotator agreement scores); and \textit{annotation information}, specifying whether additional details about annotators were provided (e.g., training they underwent, qualification criteria). We rely solely on the description of the dataset provided in the body of the accompanying paper when completing this table. }
    \label{tab:papers}
\end{table}

\begin{table}[t!]
    \scriptsize
    \centering
    \begin{tabular}{p{2.5in}p{0.3in}p{0.3in}p{0.3in}p{0.3in}p{0.3in}p{0.3in}}
    \rotatebox{0}{Dataset} & \rotatebox{45}{Standalone} & \rotatebox{45}{Concret defn.} & \rotatebox{45}{Justification} & \rotatebox{45}{Collection trade-offs} & \rotatebox{45}{Quality info} & \rotatebox{45}{Annot. info} \\
    \midrule
    LiRo~\citep{dumitrescu2021liro} &  & NA & NA &  & \checkmark & \\
    Localized Narratives~\citep{Pont_Tuset_2020} &  & \checkmark &  &  & \checkmark & \\
    MAD~\citep{Soldan_2022} &  &  &  &  & \checkmark & \checkmark\\
    Manually Annotated Sub-Corpus (MASC)~\citep{ide2008masc} & \checkmark &  &  &  & \checkmark & \\
    Mapillary Traffic Sign Dataset (MTSD)~\citep{Ertler_2020} & \checkmark & \checkmark & \checkmark &  & \checkmark & \\
    Mega-COV~\citep{Abdul_Mageed_2021} &  & \checkmark & \checkmark &  &  & NA\\
    Mickey Corpus~\citep{Lin_2021} &  & \checkmark & \checkmark & \checkmark & \checkmark & NA\\
    MightyMorph~\citep{Goldman_2021} &  & NA & NA &  &  & NA\\
    MotSynth~\citep{fabbri2021motsynth} & \checkmark &  & \checkmark & \checkmark &  & NA\\
    Multi-lingual retrieval Typologically Diverse (Mr. TyDi)~\citep{Zhang_2021} &  &  & \checkmark & \checkmark &  & \\
    MultiDoGO~\citep{Peskov_2019} &  & \checkmark &  &  & \checkmark & \checkmark\\
    Multimodal Audio-Visual Detection (MAVD)~\citep{Valverde_2021} &  & \checkmark &  &  &  & NA\\
    Natural World Tasks (NeWT)~\citep{Van_Horn_2021} & \checkmark &  & \checkmark &  & \checkmark & NA\\
    Nutrition5k~\citep{Thames_2021} &  & \checkmark & \checkmark & \checkmark &  & NA\\
    OK-VQA~\citep{Marino_2019} & \checkmark &  &  &  & \checkmark & \\
    ObjectNet~\citep{barbu2019objectnet} & \checkmark & NA & NA & \checkmark & \checkmark & NA\\
    OpoSum~\citep{Angelidis_2018} &  &  &  &  & \checkmark & \\
    PASS~\citep{asano2021pass} & \checkmark & \checkmark &  & \checkmark & \checkmark & NA\\
    Person30k~\citep{Bai_2021} &  &  & \checkmark &  &  & \\
    Query-based Video Highlights (QVHighlights)~\citep{lei2021detecting} &  &  &  &  & \checkmark & \\
    R4C~\citep{Inoue_2020} &  & NA & NA &  & \checkmark & \checkmark\\
    RID~\citep{Wan_2018} &  & \checkmark &  &  &  & NA\\
    Racial Faces in the Wild (RFW)~\citep{Wang_2019_Racial} &  &  & \checkmark &  &  & \\
    Reddit TIFU~\citep{Li_2019} &  & \checkmark &  &  & \checkmark & NA\\
    Room-Across-Room (RxR)~\citep{Ku_2020} &  & \checkmark & \checkmark & \checkmark & \checkmark & \checkmark\\
    Ruddit~\citep{Hada_2021} & \checkmark &  &  &  & \checkmark & \checkmark\\
    SCT v1.5~\citep{Sharma_2018} &  & NA & NA &  &  & NA\\
    SUNRGB-D~\citep{Song_2015} & \checkmark & NA & NA &  & \checkmark & \checkmark\\
    Salient Objects in Clutter (SOC)~\citep{Fan_2018} & \checkmark & \checkmark & \checkmark &  &  & \\
    Semantic Scholar Visual Layoutenhanced Scientific Text Understanding Evaluation
    (S2-VLUE)~\citep{Shen_2022} &  &  & \checkmark &  &  & \\
    SketchyScene~\citep{Zou_2018} &  & \checkmark &  &  & \checkmark & \\
    StackEx~\citep{Yuan_2020} &  &  &  & \checkmark & \checkmark & NA\\
    StockEmotion~\citep{wei2020learning} &  & \checkmark & \checkmark &  & \checkmark & \\
    SynthBio~\citep{yuan2021synthbio} &  & NA & NA & \checkmark & \checkmark & \checkmark\\
    Synthia~\citep{Ros_2016} & \checkmark &  &  & \checkmark &  & NA\\
    TVR~\citep{Lei_2020_TVR} &  & \checkmark & \checkmark &  & \checkmark & \checkmark\\
    Taskmaster-1~\citep{Byrne_2019} &  &  &  & \checkmark & \checkmark & \checkmark\\
    The Dollar Street Dataset~\citep{rojas2022dollar} & \checkmark & \checkmark & \checkmark & \checkmark & \checkmark & NA\\
    TicketTalk~\citep{Byrne_2021} &  &  & \checkmark &  &  & \\
    Tracking Any Object (TAO)~\citep{Dave_2020} & \checkmark &  & \checkmark &  & \checkmark & \\
    TrackingNet~\citep{M_ller_2018} & \checkmark &  &  &  & \checkmark & \\
    Trans10K~\citep{Xie_2020} &  & \checkmark & \checkmark &  &  & \\
    UCF-QNRF~\citep{Idrees_2018} &  & \checkmark &  &  & \checkmark & \\
    VCR~\citep{Zellers_2019} &  &  &  & \checkmark & \checkmark & \checkmark\\
    VIdeo Panoptic Segmentation in the Wild (VIPSeg)~\citep{Miao_2022} &  & \checkmark & \checkmark & \checkmark & \checkmark & \\
    VPS~\citep{Miao_2021} &  &  &  & \checkmark &  & \\
    VTUAV~\citep{Zhang_2022} &  & \checkmark &  &  & \checkmark & NA\\
    Vehicle Re-Identification for Aerial Image (VRAI)~\citep{Wang_2019} &  & \checkmark & \checkmark &  &  & \\
    \bottomrule
    \end{tabular}
    \caption{(Continued from~\Cref{tab:papers}). Summary of our analysis across all 135 image and text datasets. Column labels: \textit{standalone}, indicating whether the dataset paper exclusively presents a dataset without introducing novel models, tasks, or algorithms; \textit{concrete definition}, indicating whether the dataset paper includes a clear definition of diversity (\emph{NA} for datasets focusing solely on bias, not diversity); \textit{justification}, denoting whether the dataset paper discusses the rationale behind incorporating diversity into the dataset (\emph{NA} for datasets focusing solely on bias, not diversity); \textit{collection trade-offs}, indicating whether the dataset paper discusses drawbacks of the chosen collection methodology or provides considerations for alternative methodologies; \textit{quality information}, signifying whether the dataset paper includes details on how dataset quality was verified (e.g., manual inspection by creators, inter-annotator agreement scores); and \textit{annotation information}, specifying whether additional details about annotators were provided (e.g., training they underwent, qualification criteria). We rely solely on the description of the dataset provided in the body of the accompanying paper when completing this table. }
    \label{tab:papers2}
\end{table}
\begin{table}[]
    \scriptsize
    \centering
    \begin{tabular}{p{2.5in}p{0.3in}p{0.3in}p{0.3in}p{0.3in}p{0.3in}p{0.3in}}
    \rotatebox{0}{Dataset} & \rotatebox{45}{Standalone} & \rotatebox{45}{Concret defn.} & \rotatebox{45}{Justification} & \rotatebox{45}{Collection trade-offs} & \rotatebox{45}{Quality info} & \rotatebox{45}{Annot. info} \\
    \midrule
    VidSitu~\citep{Sadhu_2021} & \checkmark &  &  &  & \checkmark & \checkmark\\
    Video gAze CommunicATION (VACATION)~\citep{Fan_2019_gaze} &  & \checkmark & \checkmark &  & \checkmark & \\
    Video-and-Language Event Prediction (VLEP)~\citep{Lei_2020} &  & \checkmark &  &  & \checkmark & \\
    VideoCoAtt~\citep{Fan_2018} &  & \checkmark & \checkmark &  &  & \\
    VisDrone-DET 2018~\citep{Cao_2021} &  & \checkmark &  &  &  & \\
    Vision-based Fallen Person (VFP290K)~\citep{an2021vfp290k} &  \checkmark & \checkmark & \checkmark &  & \checkmark & \checkmark\\
    WIDERFACE-DEMO~\citep{Yang_2022} &  &  & \checkmark &  &  & \\
    WIKIBIAS~\citep{Zhong_2021} &  & NA & NA &  & \checkmark & \\
    Waymo Open Dataset~\citep{Sun_2020} & \checkmark & \checkmark & \checkmark &  & \checkmark & \\
    Waymo Open Motion Dataset~\citep{ettinger2021large} &  &  & \checkmark &  &  & NA\\
    WikiEvolve~\citep{De_Kock_2022} &  & NA & NA &  &  & NA\\
    Wilddash2 (WD2)~\citep{Zendel_2022} &  & \checkmark &  & \checkmark &  & \\
    Winobias~\citep{Zhao_2018} &  & NA & NA &  &  & NA\\
    Winogender~\citep{Rudinger_2018} &  & NA & NA &  &  & \\
    X-CSQA~\citep{Lin_2021} &  & \checkmark & \checkmark & \checkmark &  & NA\\
    YASO~\citep{Orbach_2021} & \checkmark &  & \checkmark & \checkmark & \checkmark & \\
   \citet{Culkin_2021} &  &  &  &  & \checkmark & \\
    \citet{Drawzeski_2021} &  &  & \checkmark & \checkmark &  & \\
    \citet{Durmus_2019} &  & \checkmark & \checkmark &  &  & \\
    \citet{Fu_2021} &  & \checkmark & \checkmark &  &  & NA\\
    \citet{Gillani_2019} &  & NA & NA &  &  & NA\\
    \citet{Johnson_2018} &  & NA & NA &  & \checkmark & \checkmark\\
    \citet{Lepp_2020} &  & NA & NA &  & \checkmark & \checkmark\\
    \citet{Ni_2019} &  & NA & NA &  &  & NA\\
    \citet{Pratapa_2022} &  & \checkmark &  &  &  & NA\\
    \citet{Schulz_2019} &  & NA & NA &  & \checkmark & \checkmark\\
    \citet{Silberman_2012} &  &  &  &  &  & \\
    \citet{Sugawara_2022} & \checkmark &  & \checkmark &  & \checkmark & \checkmark\\
    \citet{Sun_2021} &  &  & \checkmark &  & \checkmark & \\
    \citet{Wang_2019} &  & \checkmark & \checkmark &  &  & NA\\
    \citet{Wiegand_2021} &  & NA & NA &  & \checkmark & \checkmark\\
    \citet{Zhang_2020} &  &  &  &  &  & NA\\
    \citet{Zhao_2021} &  & NA & NA & \checkmark & \checkmark & \checkmark\\
    \citet{celis2016fair} &  & \checkmark & \checkmark &  &  & NA\\
    \citet{hemani2021ails} &  & NA & NA &  &  & NA\\
    \citet{rahman2021information} &  &  & \checkmark &  & \checkmark & \checkmark\\
    iEAT~\citep{Steed_2021} &  & NA & NA &  &  & NA\\
    iNat2021~\citep{Van_Horn_2021} & \checkmark & \checkmark & \checkmark & \checkmark &  & \\
    iVQA~\citep{Yang_2021} &  &  &  &  & \checkmark & \\
    mTVR~\citep{Lei_2021} &  & \checkmark &  &  & \checkmark & \checkmark\\
    \bottomrule
    \end{tabular}
    \caption{(Continued from~\Cref{tab:papers2}). Summary of our analysis across all 135 image and text datasets. Column labels: \textit{standalone}, indicating whether the dataset paper exclusively presents a dataset without introducing novel models, tasks, or algorithms; \textit{concrete definition}, indicating whether the dataset paper includes a clear definition of diversity (\emph{NA} for datasets focusing solely on bias, not diversity); \textit{justification}, denoting whether the dataset paper discusses the rationale behind incorporating diversity into the dataset (\emph{NA} for datasets focusing solely on bias, not diversity); \textit{collection trade-offs}, indicating whether the dataset paper discusses drawbacks of the chosen collection methodology or provides considerations for alternative methodologies; \textit{quality information}, signifying whether the dataset paper includes details on how dataset quality was verified (e.g., manual inspection by creators, inter-annotator agreement scores); and \textit{annotation information}, specifying whether additional details about annotators were provided (e.g., training they underwent, qualification criteria). We rely solely on the description of the dataset provided in the body of the accompanying paper when completing this table.}
    \label{tab:papers3}
\end{table}
\newpage

\end{document}